\documentclass{article}
\usepackage{iclr2026_conference,times}

\usepackage{amsmath,amsfonts,bm}

\def\eqref#1{equation~\ref{#1}}

\def\1{\bm{1}}

\DeclareMathAlphabet{\mathsfit}{\encodingdefault}{\sfdefault}{m}{sl}
\SetMathAlphabet{\mathsfit}{bold}{\encodingdefault}{\sfdefault}{bx}{n}

\usepackage{hyperref}
\usepackage{url}
\usepackage{amsmath,amsfonts}
\usepackage{graphicx}
\usepackage{booktabs}
\usepackage{arydshln}
\usepackage{subcaption}
\usepackage[table]{xcolor}
\usepackage{pgf}
\usepackage{multirow}
\usepackage[T1]{fontenc}
\usepackage[utf8]{inputenc}
\usepackage{xcolor}
\usepackage{listings}
\usepackage{listingsutf8}
\usepackage{float}

\definecolor{vanillacolor}{RGB}{240,240,240}
\definecolor{easyocrcolor}{RGB}{255,245,230}
\definecolor{mistralcolor}{RGB}{230,245,255}
\definecolor{vlmcolor}{RGB}{230,255,230}

\newcommand{\example}[3]{%
    \subsection{#1}
    \begin{minipage}[t]{0.35\textwidth}
    \vspace{0pt}
    \includegraphics[width=\textwidth]{images/#2}
    \end{minipage}%
    \hfill
    \begin{minipage}[t]{0.62\textwidth}
    \vspace{0pt}
    {\scriptsize\textbf{Adobe Text Extract}}
    \lstinputlisting[backgroundcolor=\color{vanillacolor}]{annotations/#3/vanilla.txt}
    {\scriptsize\textbf{EasyOCR}}
    \lstinputlisting[backgroundcolor=\color{easyocrcolor}]{annotations/#3/easyocr.txt}
    \end{minipage}
    \vspace{0.3em}
    \begin{minipage}[t]{0.48\textwidth}
    {\scriptsize\textbf{Mistral OCR 3}}
    \lstinputlisting[backgroundcolor=\color{mistralcolor}]{annotations/#3/mistral_cleaned.txt}
    \end{minipage}%
    \hfill
    \begin{minipage}[t]{0.48\textwidth}
    {\scriptsize\textbf{Ministral 3B}}
    \lstinputlisting[backgroundcolor=\color{vlmcolor}]{annotations/#3/vlm.txt}
    \end{minipage}
    \vspace{1em}
    \hrule
    \vspace{1em}
}

\usepackage[table]{xcolor}   
\usepackage{colortbl}        

\definecolor{GoldBg}{RGB}{255, 221, 102}
\definecolor{SilverBg}{RGB}{210, 210, 210}
\definecolor{BronzeBg}{RGB}{214, 170, 120}

\title{Retrieval or Representation? Reassessing Benchmark Gaps in Multilingual and Visually Rich RAG}

\author{
  \begin{tabular}[t]{@{}l@{\hspace{1.2cm}}l@{}}
    \begin{tabular}[t]{@{}l@{}}
      \textbf{Martin Asenov} \\
      {\normalfont Parexel AI Labs} \\
      {\normalfont London, United Kingdom} \\
      {\normalfont \texttt{martin.asenov@parexel.com}}
    \end{tabular}
    &
    \begin{tabular}[t]{@{}l@{}}
      \textbf{Kenza Benkirane} \\
      {\normalfont Parexel AI Labs} \\
      {\normalfont London, United Kingdom} \\
      {\normalfont \texttt{kenza.benkirane@parexel.com}}
    \end{tabular}
    \\[4em]
    \begin{tabular}[t]{@{}l@{}}
      \textbf{Dan Goldwater} \\
      {\normalfont Parexel AI Labs} \\
      {\normalfont London, United Kingdom} \\
      {\normalfont \texttt{dan.goldwater@parexel.com}}
    \end{tabular}
    &
    \begin{tabular}[t]{@{}l@{}}
      \textbf{Aneiss Ghodsi} \\
      {\normalfont Parexel AI Labs} \\
      {\normalfont San Francisco, United-States} \\
      {\normalfont \texttt{aneiss.ghodsi@parexel.com}}
    \end{tabular}
  \end{tabular}
}
\iclrfinalcopy
\begin{document}
\maketitle

\begin{abstract}


Retrieval-augmented generation (RAG) is a common way to ground language models in external documents and up-to-date information. Classical retrieval systems relied on lexical methods such as BM25, which rank documents by term overlap with corpus-level weighting. End-to-end multimodal retrievers trained on large query-document datasets claim substantial improvements over these approaches, especially for multilingual documents with complex visual layouts. We demonstrate that better document representation is the primary driver of benchmark improvements. By systematically varying transcription and preprocessing
methods while holding the retrieval mechanism fixed, we demonstrate that BM25 can recover large gaps on multilingual and visual benchmarks. Our findings call for decomposed evaluation benchmarks that separately measure transcription and retrieval capabilities, enabling the field to correctly attribute progress and focus effort where it matters.

\end{abstract}

\begin{figure*}[h]
  \centering
  \includegraphics[width=\textwidth]{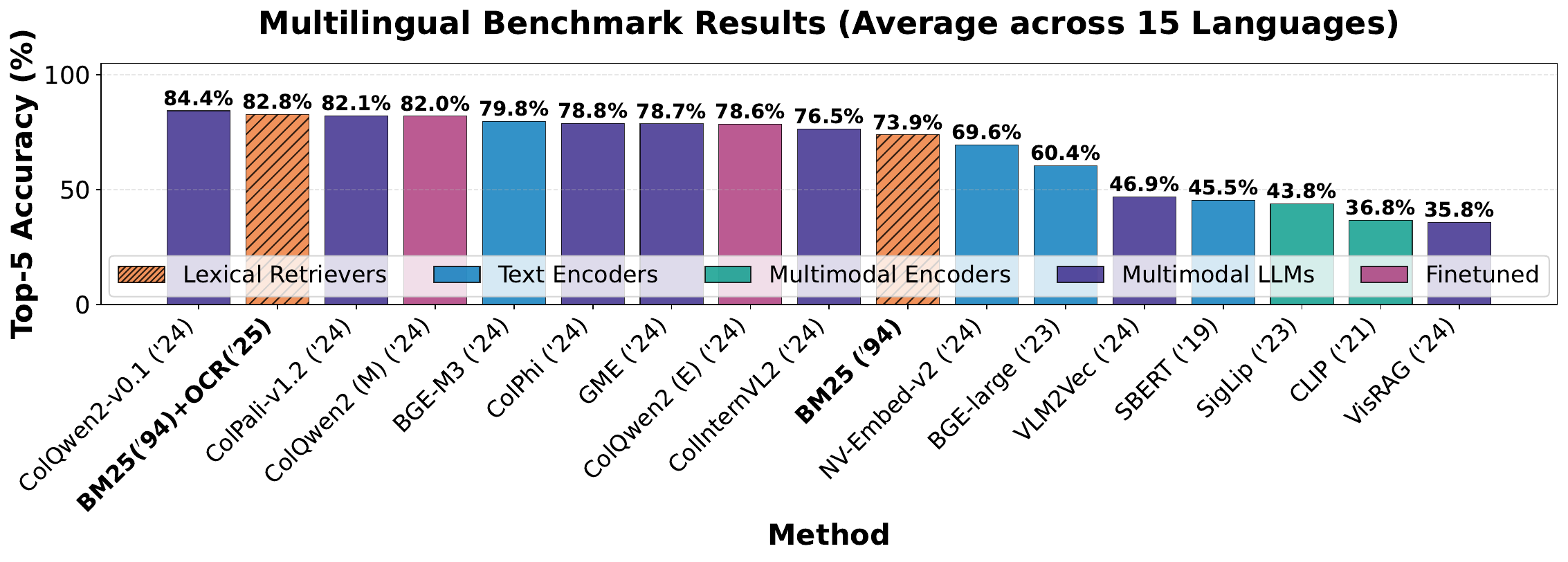}
  \caption{\textbf{Multilingual benchmark results across 15 languages.}
  Average Top-5 retrieval accuracy for different methods. Methods are sorted by performance (highest to lowest).
  Lexical retrievers (BM25) are shown with diagonal hatching.
  BM25+OCR indexes text produced by state-of-the-art OCR models and preprocessing techniques per different languages.
  Release years are shown in parentheses.}
  \label{fig:multilingual_results}
\end{figure*}
\vspace{-2pt}

\section{Introduction}

Despite recent progress, retrieval in multilingual and visually rich settings remains challenging for modern systems. Multilingual benchmarks and training corpora are heavily skewed toward high-resource languages, leading to persistent performance gaps and the need for per-language optimizations~\cite{chirkova2024retrieval,ranaldi2025multilingual,li2025language}. Many real-world documents interleave running text with figures, tables, and complex layouts, introducing additional challenges for document retrieval~\cite{tanaka2025vdocrag}. Recent work has increasingly emphasized specialized retrievers, including dense text embeddings~\cite{xiao2023cpack,chen2024bgem3}, multimodal representations~\cite{radford2021clip,zhai2023siglip,faysse2024colpali}, or layout-aware models~\cite{layoutlm,layoutlmv2,layoutlmv3}, to address the perceived shortcomings of classical text retrieval methods such as BM25~\cite{robertson1994some}. Empirical results on multimodal, visually rich benchmarks such as VisR-Bench~\cite{chen2025visr} appear to support this trend, showing large performance gaps between sparse text retrievers and modern multimodal approaches.

To examine the impact of optical character recognition (OCR) and text preprocessing, we extend transcription data in VisR-Bench~\cite{chen2025visr} with three additional OCR models and implement language-specific preprocessing options, including stemming, lemmatization, and morphological analysis. Our results show that improving transcription quality and the resulting text representations leads to significantly better downstream retrieval performance - recovering up to +8.9 Top-5 points for BM25~\cite{robertson1994some} on average for multilingual datasets by improving transcription quality and normalization. On figure-heavy pages, we observe a distinct failure mode: when figures lack any textual or semantic description, retrieval degrades sharply, whereas even coarse descriptions recover much of the loss, yielding gains of up to +31.1 Top-5 points. This raises a basic question: \emph{are we evaluating retrieval methods, or the pre-processing pipelines?} We show that addressing these sources of error alone allows classical methods such as BM25 to recover a large fraction of the apparent gap.

\section{Related work and preliminaries}

\textbf{Multilingual retrieval} performance remains uneven because benchmarks and training data are skewed toward high-resource languages, leaving persistent gaps in low-resource and morphologically rich settings~\cite{chirkova2024retrieval,ranaldi2025multilingual,li2025language}. Recent work therefore emphasizes multilingual dense retrievers and embedding models to improve semantic matching across languages~\cite{xiao2023cpack,chen2024bgem3}.  Any text retriever is fundamentally bounded by the indexed representation, which depends on OCR quality and language-specific preprocessing such as tokenization, stemming, and morphological normalization. Our work complements model-centric studies by showing that much of the observed multilingual gap can be driven by these representation choices rather than retrieval alone.

\textbf{Visually rich documents} interleave text with figures and complex layouts, motivating layout-aware encoders and multimodal retrievers that bypass brittle text extraction~\cite{tanaka2025vdocrag,layoutlm,layoutlmv2,layoutlmv3,radford2021clip,zhai2023siglip,faysse2024colpali}. Benchmarks such as VisR-Bench~\cite{chen2025visr} report large gains for multimodal methods over classical sparse retrieval methods such as BM25~\cite{robertson1994some}. A key caveat is that these comparisons often confound retrieval with upstream transcription quality: if figure content is poorly transcribed, it is effectively missing from the text index. We address this by holding the retriever fixed and varying only OCR/transcription, isolating how improved extraction (and lightweight figure descriptions) can substantially narrow the apparent gap.

\section{Experiments}
\label{sec:experiments}
\subsection{Experimental setup}
\textbf{Benchmark and metrics} Our goal is to separate retrieval behavior from upstream transcription and normalization choices. We run controlled experiments where we keep the retriever and evaluation protocol fixed while varying only (i) the OCR/transcription used to build the page index, and (ii) language-specific text processing for multilingual retrieval. We evaluate on VisR-Bench~\cite{chen2025visr}, a benchmark for retrieval-augmented question answering over long, visually rich documents. Each example contains a document, a query, and a ground-truth evidence page. The task is \emph{page-level retrieval}: return the evidence page in the Top-$K$ retrieved pages. We report results (i) across 15 languages in Figure~\ref{fig:multilingual_results}, and (ii) across visually specific questions in FIgure~\ref{fig:figure_results}.

\textbf{Retrievers} We compare three classes of retrieval methods: (i) sparse lexical retrieval using BM25~\cite{robertson1994some}; (ii) dense text retrievers including SBERT~\cite{reimers2019sentencebert}, BGE-large~\cite{xiao2023cpack}, BGE-M3~\cite{chen2024bgem3}, and NV-Embed-v2~\cite{lee2024nvembed}; and (iii) multimodal retrievers including CLIP~\cite{radford2021clip}, SigLip~\cite{zhai2023siglip}, VisRAG~\cite{yu2024visrag}, VLM2Vec~\cite{jiang2024vlm2vec}, GME~\cite{zhang2024gme}, and Col* methods~\cite{chen2024sdrag,faysse2024colpali}. We report baseline results from~\cite{chen2025visr} for multimodal models, while our evaluation focuses on varying OCR and preprocessing methods. Text-based retrievers index OCR transcriptions of each page, while multimodal methods encode page images directly.

\begin{figure*}[t]
\centering
\begin{subfigure}[b]{0.48\textwidth}
  \includegraphics[width=\textwidth]{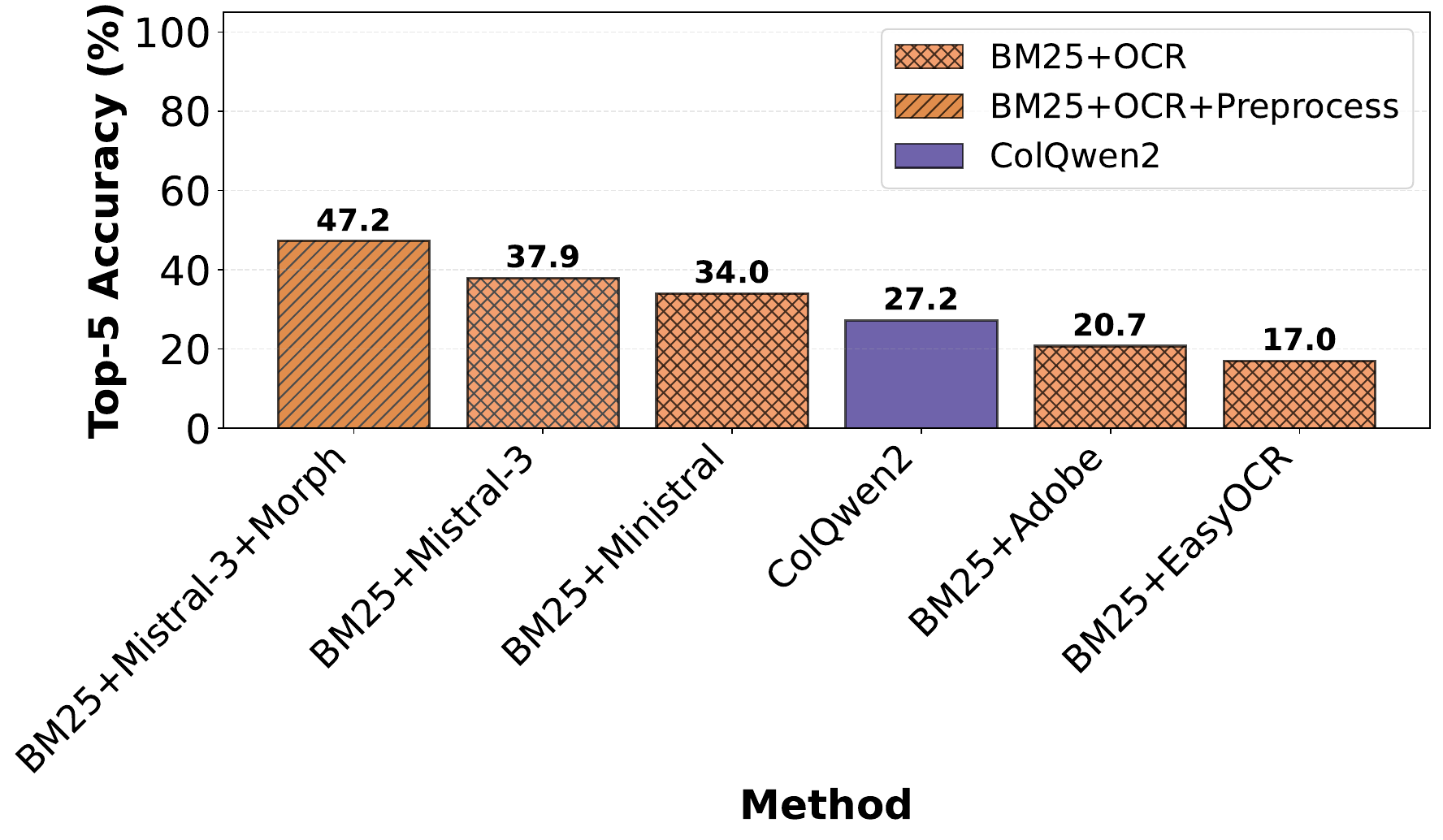}
  \caption{Arabic}
\end{subfigure}
\hfill
\begin{subfigure}[b]{0.48\textwidth}
  \includegraphics[width=\textwidth]{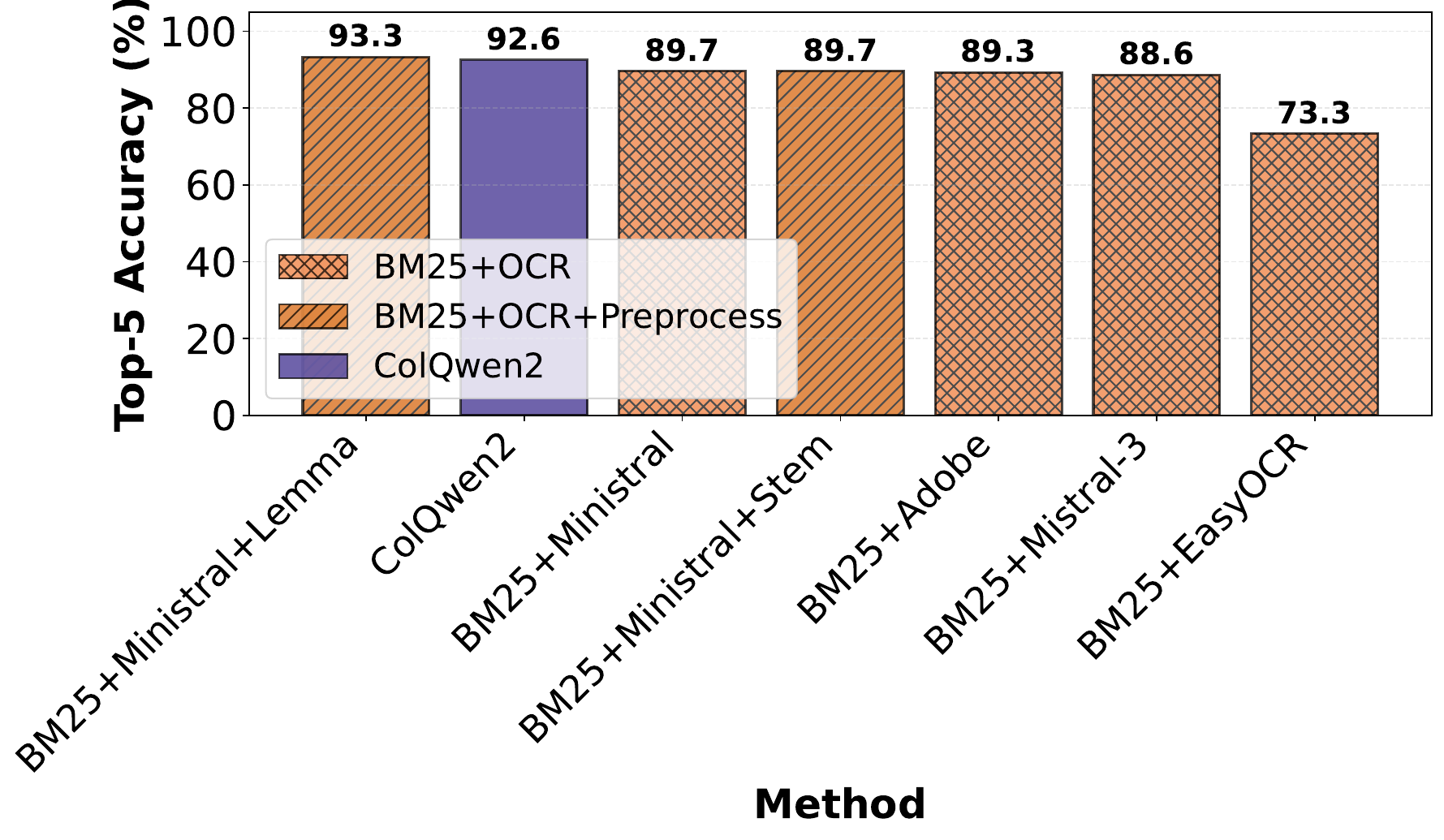}
  \caption{Czech}
\end{subfigure}

\begin{subfigure}[b]{0.48\textwidth}
  \includegraphics[width=\textwidth]{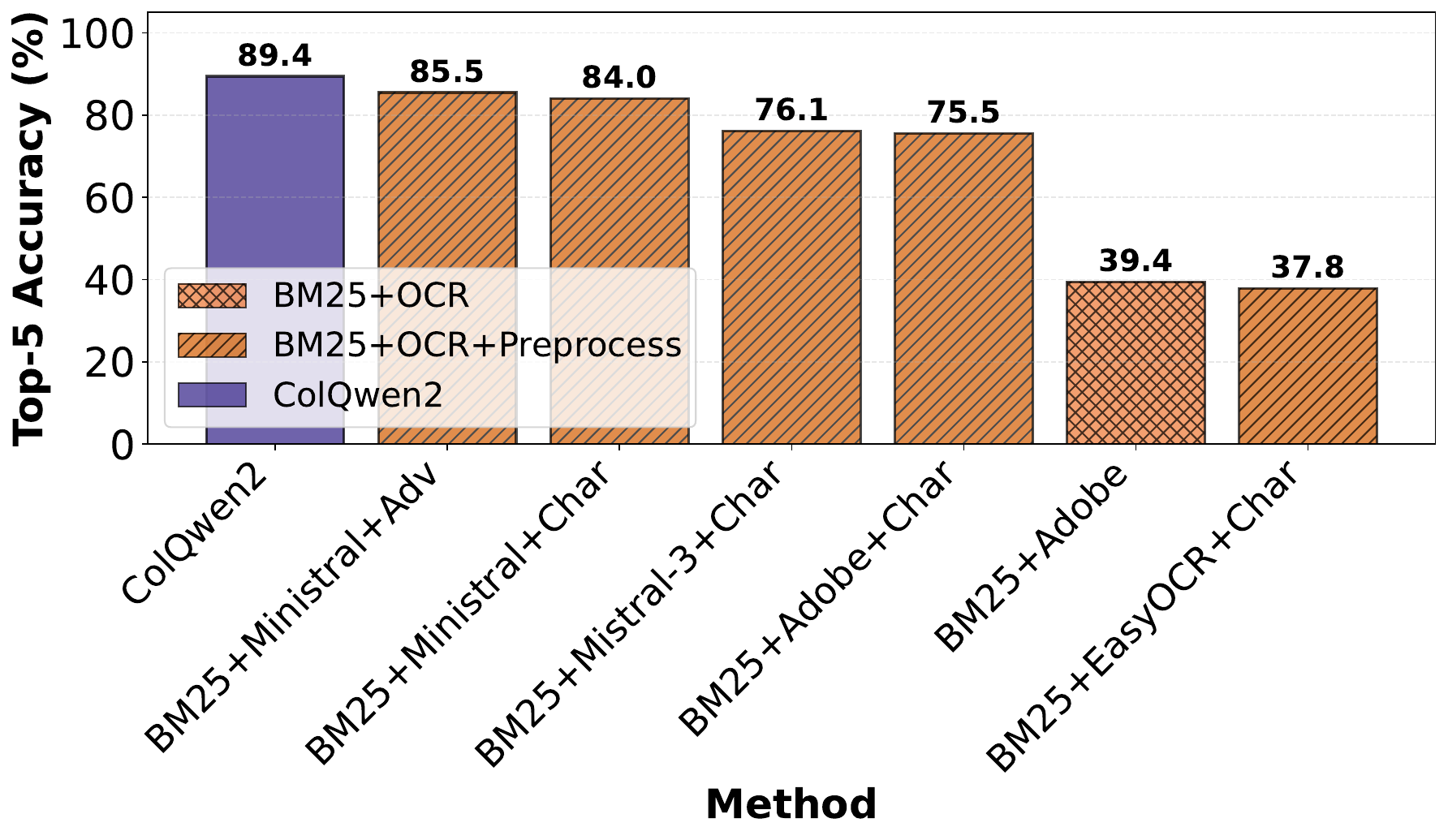}
  \caption{Japanese}
\end{subfigure}
\hfill
\begin{subfigure}[b]{0.48\textwidth}
  \includegraphics[width=\textwidth]{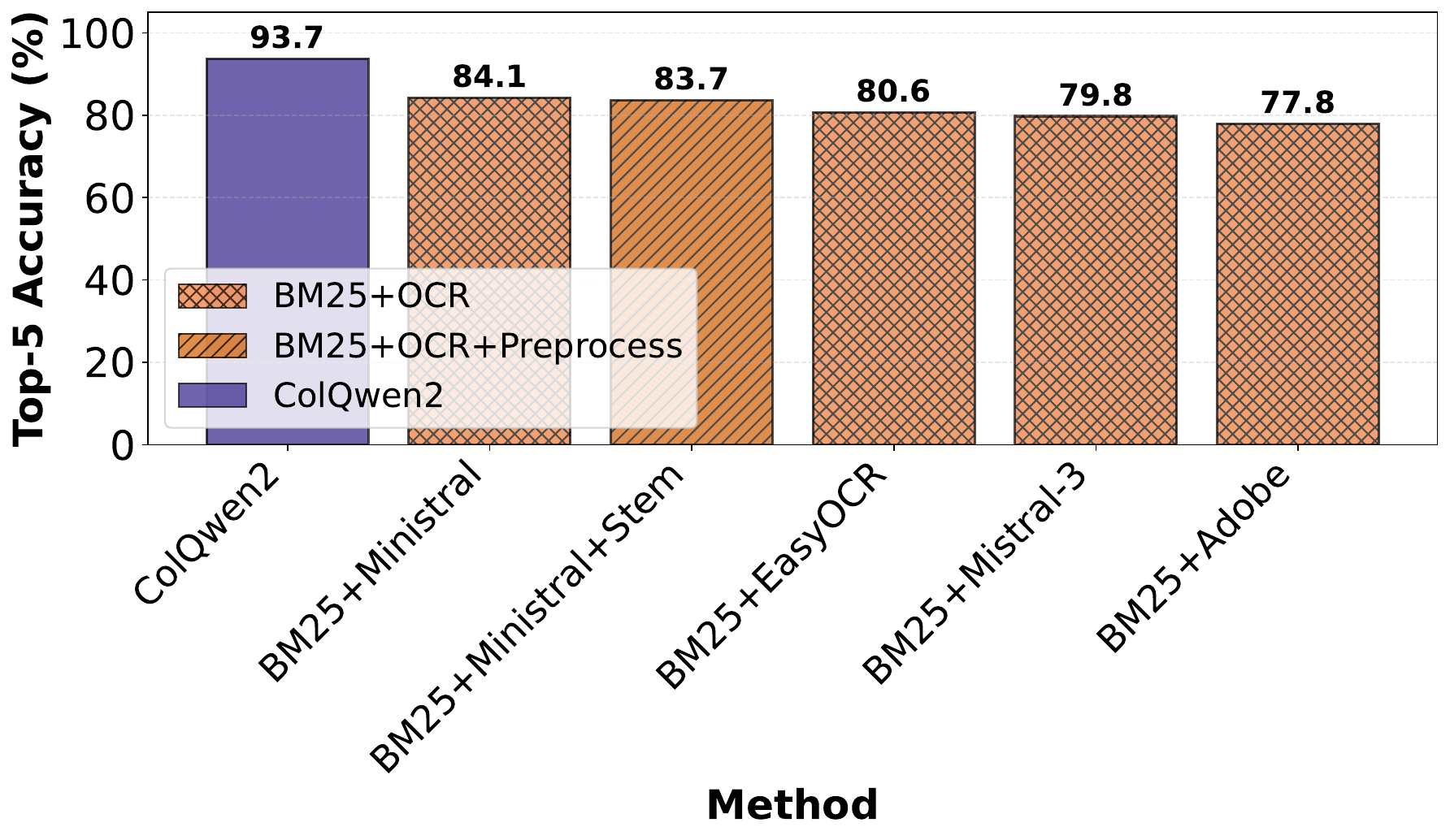}
  \caption{French}
\end{subfigure}

\caption{\textbf{Language-specific BM25 optimization.}
Top-5 retrieval accuracy for BM25 under different OCR/transcription pipelines (Adobe, EasyOCR, Ministral 3B, Mistral OCR 3) and language-specific text processing (lemmatization, stemming, morphological analysis, and segmentation).}
\label{fig:bm25_configurations}
\end{figure*}
\vspace{-2pt}

\textbf{OCR models} A central variable in our study is transcription quality. We compare the dataset-provided extraction against alternative OCR pipelines:

\begin{itemize}
    \item \textbf{Adobe Document Extract}: default parser in the dataset~\cite{adobe_pdf_extract_api}
    \item \textbf{EasyOCR}: open-source OCR applied to rendered page images~\cite{easyocr2020}.
    \item \textbf{Mistral OCR 3}: a modern OCR system applied to page images~\cite{mistral_ocr_3_2025}.
    \item \textbf{Ministral 3B (VLM transcription)}: a small VLM~\cite{ministral} with the prompt: \emph{``Give me a markdown of what you see in the image. Reply only with the markdown content.''}
\end{itemize}
Adobe Document Extract and EasyOCR perform text-only extraction, while Mistral OCR 3 annotates figures and tables. Ministral 3B is applied per image with the specified prompt.

\textbf{Language-Specific Text Processing} For multilingual BM25, we evaluate a set of language-specific preprocessing strategies that directly affect lexical matching. We consider stemming for Romance and Germanic languages using Snowball stemmers~\cite{porter2001snowball}; lemmatization for highly inflected languages such as Czech, Slovenian, Croatian, and Finnish using spaCy models~\cite{honnibal2020spacy}; full morphological analysis for Arabic using CAMeL Tools~\cite{obeid2020camel}, which decomposes words into prefixes, stems, and suffixes; and script-aware word segmentation for Japanese with Sudachi~\cite{takaoka2018sudachi} and for Vietnamese with pyvi~\cite{pyvi}. As a reference, we include a minimal-processing baseline that applies only lowercasing and NLTK tokenization~\cite{bird2006nltk}. For each language, we select the best-performing configuration based on Top-5 accuracy and report the chosen setup in Figure~\ref{fig:multilingual_results}, with configuration sensitivity illustrated for representative languages in Figure~\ref{fig:bm25_configurations}.

\subsection{Results}
\label{sec:results_overview}

\begin{figure*}[h]
      \centering
      \begin{subfigure}[b]{0.49\textwidth}
          \centering
          \includegraphics[width=\textwidth]{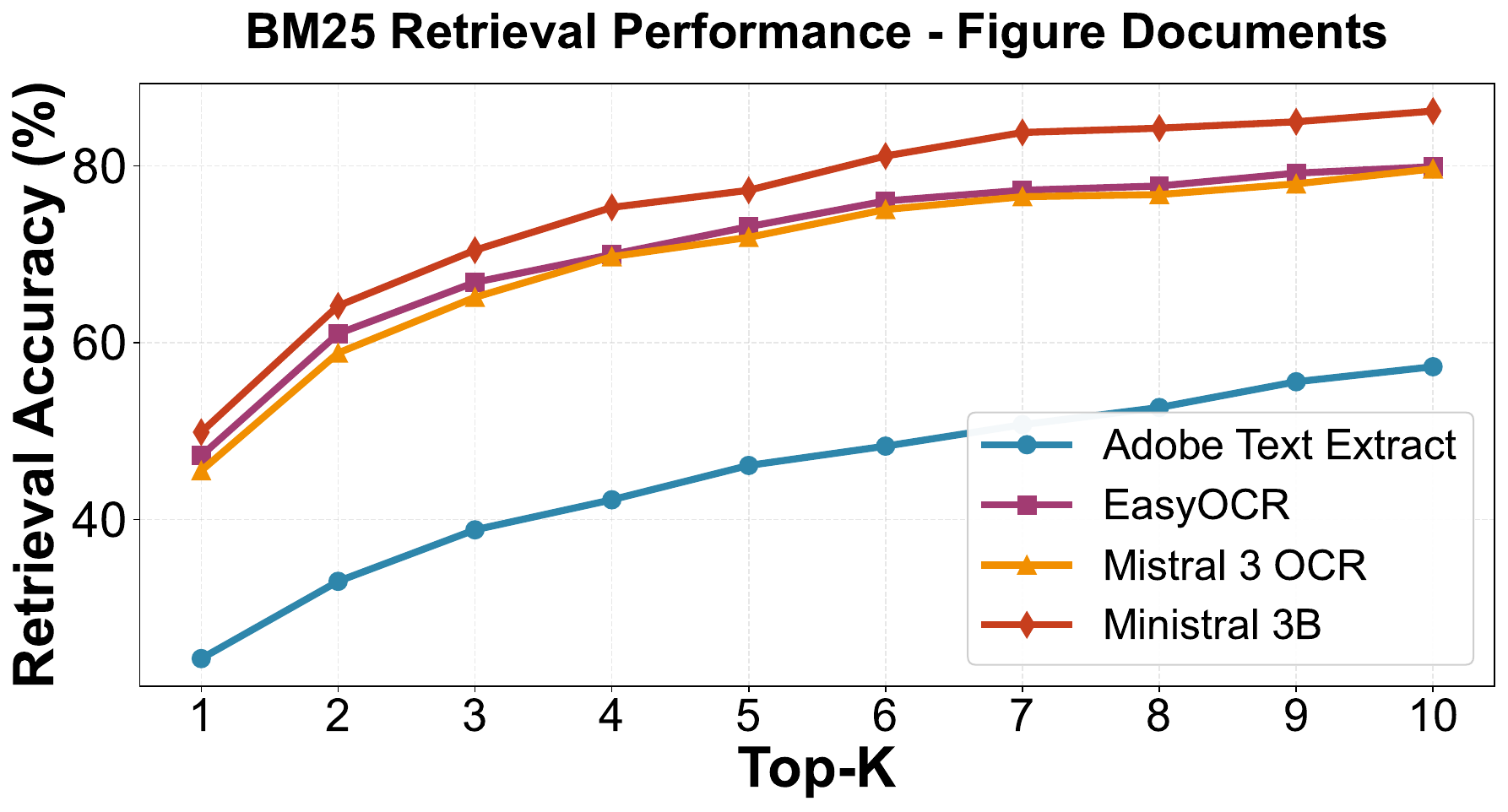}
          \caption{Figure Documents}
          \label{fig:bm25_figure}
      \end{subfigure}
      \hfill
      \begin{subfigure}[b]{0.49\textwidth}
          \centering
          \includegraphics[width=\textwidth]{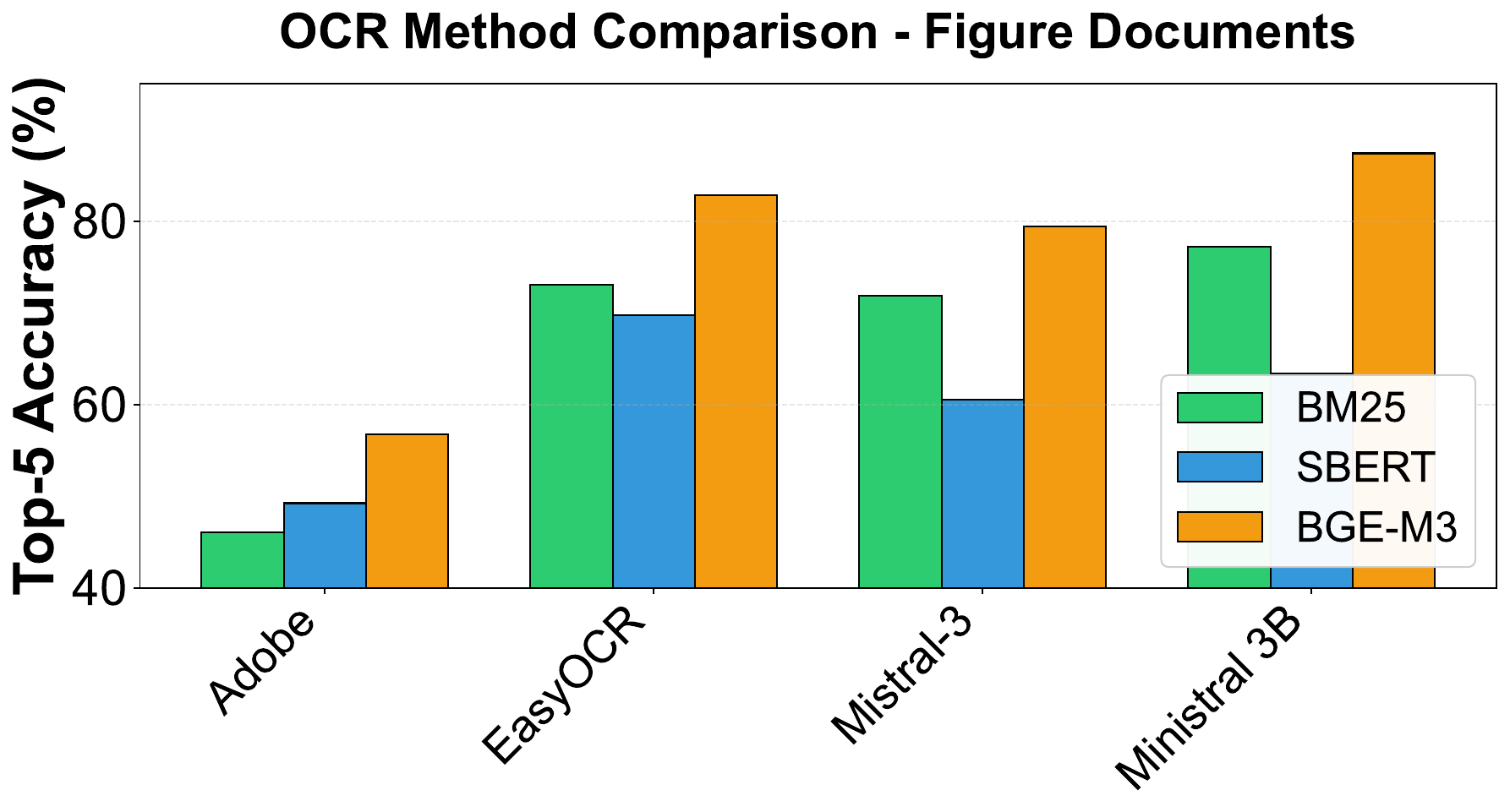}
          \caption{Figure Documents}
          \label{fig:ocr_figure}
      \end{subfigure}
      \caption{\textbf{OCR impact on text retrieval methods.} a) BM25 retrieval performance for different Top-K values b) Impact on different retrieval for different combinations of transcription OCR models and text retrieval models.}
      \label{fig:bm25_comparison}
  \end{figure*}
\vspace{-1pt}

\begin{figure*}[h]
  \centering
  \includegraphics[width=\textwidth]{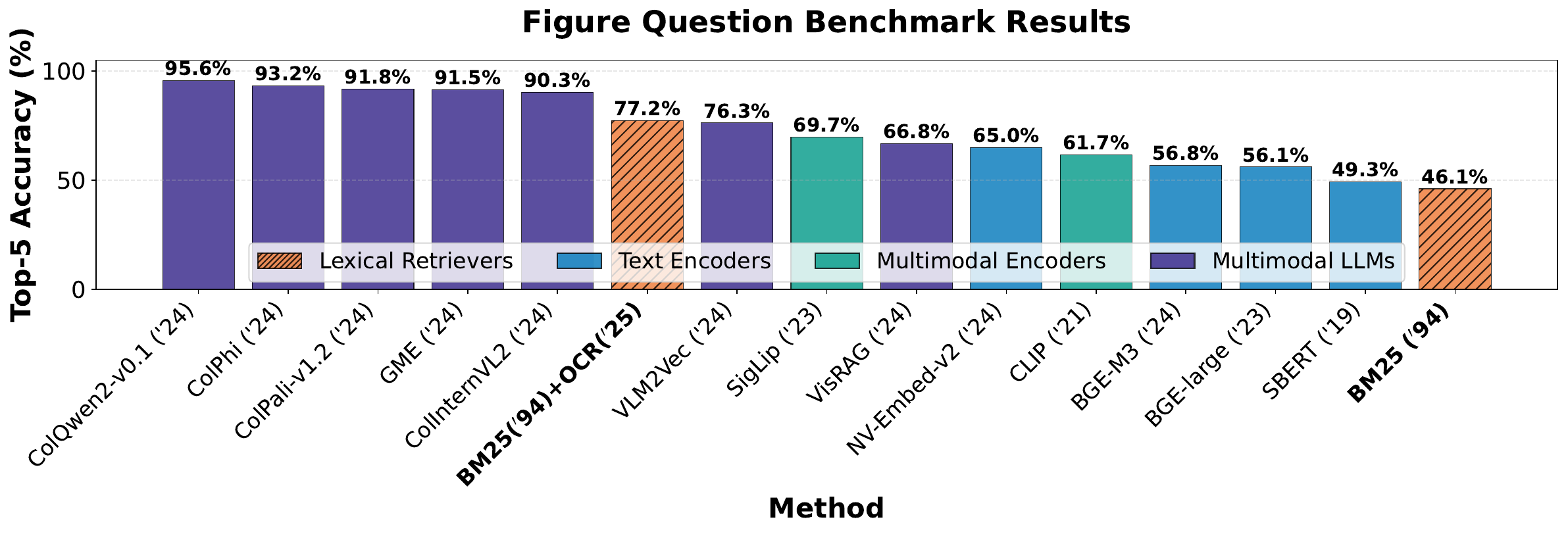}
  \caption{\textbf{Figure-heavy focused QA benchmark results.}
  Average Top-5 retrieval accuracy for different methods. Methods are sorted by performance (highest to lowest).
  Lexical retrievers (BM25) are shown with diagonal hatching.
  BM25+OCR uses a small visual language model~\cite{ministral} for transcription.
  Release years are shown in parentheses.}
  \label{fig:figure_results}
\end{figure*}
\vspace{-2pt}

\textbf{Multilingual} Across multilingual settings, we observe that retrieval performance is strongly shaped by transcription and preprocessing choices (Figure~\ref{fig:multilingual_results}). Modern OCR substantially improves baseline accuracy, and languages with complex morphology or segmentation (e.g., Japanese, Arabic) benefit disproportionately from appropriate tokenization and morphological processing (Figure~\ref{fig:bm25_configurations}). 

\textbf{Visually Rich} 
For figure-heavy pages, transcribing visual content and adding semantic descriptions recovers most of the performance gap (Figure~\ref{fig:figure_results}). We observe a two-stage progression: gains from improved transcription fidelity, followed by additional improvements from semantic figure descriptions (Figure~\ref{fig:bm25_figure}). These improvements benefit not only BM25 but also dense text retrievers such as SBERT and BGE (Figure~\ref{fig:ocr_figure}).

\textbf{Across Both Settings} Across both multilingual and visually rich retrieval, these results indicate that missing or noisy transcription is a dominant failure mode (Figure~\ref{fig:multilingual_results}, Figure~\ref{fig:figure_results}). Retrieval outcomes are therefore strongly shaped by upstream extraction and processing.

\vspace{-1pt}
\section{Conclusion}
\label{sec:conclusion}

  We revisit the narrative that lexical retrieval underperforms on visually rich and multilingual benchmarks due to inadequate text matching. We show that OCR quality is a critical bottleneck: improving transcription alone recovers most of the performance gap previously attributed to retrieval limitations. These findings call for treating OCR as a first-class component of document retrieval systems.


\bibliography{ocr_ref}

\clearpage
\appendix


\clearpage
\appendix

\section{Usage of LLM models disclosure}
Large language models were used for proofreading and rephrasing individual sentences of the manuscript for clarity and conciseness. All claims and results were independently produced and validated by the authors.
\section{Multilingual Retrieval: OCR vs.\ Preprocessing}
\label{app:multilingual_findings}

This section comments on multilingual patterns that are visible in the full ablations and per-language
config summaries (Table~\ref{tab:bm25_complete}, Table~\ref{tab:best_configs}) and the full multilingual
leaderboard (Table~\ref{tab:multilingual_full}), but are not discussed in the main text.

\subsection{When OCR Dominates vs.\ When Preprocessing Dominates}
\label{app:ocr_vs_norm}

The complete BM25 ablation in Table~\ref{tab:bm25_complete} reveals two regimes.

\paragraph{OCR-dominated languages.}
For several languages (e.g., Arabic, Japanese, Vietnamese), the choice of OCR model explains most of the
performance variance. In these cases, linguistic normalization cannot compensate for missing or corrupted
transcription; the dominant gains come from recovering readable text (Table~\ref{tab:bm25_complete}).

\paragraph{Preprocessing-dominated languages.}
For highly inflected languages (e.g., Czech, Slovenian, Croatian), OCR choice is often secondary once a
reasonable transcription is available. Here, lemmatization or morphology accounts for most of the gains,
suggesting that representation quality is limited less by OCR and more by token normalization
(Table~\ref{tab:bm25_complete}, Table~\ref{tab:best_configs}).

These regimes help explain why a single global preprocessing pipeline is suboptimal (Table~\ref{tab:best_configs}).

\begin{table*}[h]
\centering
\caption{\textbf{Multilingual retrieval accuracy (\%) on VisR-Bench.}
Top-1 and Top-5 accuracy across 15 languages.}
\label{tab:multilingual_full}
\resizebox{\textwidth}{!}{%
\begin{tabular}{l|cc|cc|cc|cc|cc|cc|cc|cc}
\toprule

& \multicolumn{2}{c}{\textbf{Spanish}}
& \multicolumn{2}{c}{\textbf{Italian}}
& \multicolumn{2}{c}{\textbf{German}}
& \multicolumn{2}{c}{\textbf{French}}
& \multicolumn{2}{c}{\textbf{Dutch}}
& \multicolumn{2}{c}{\textbf{Arabic}}
& \multicolumn{2}{c}{\textbf{Croatian}}
& \multicolumn{2}{c}{\textbf{Japanese}} \\

\cmidrule(lr){2-3}\cmidrule(lr){4-5}\cmidrule(lr){6-7}\cmidrule(lr){8-9}
\cmidrule(lr){10-11}\cmidrule(lr){12-13}\cmidrule(lr){14-15}\cmidrule(lr){16-17}

Method
& top1 & top5
& top1 & top5
& top1 & top5
& top1 & top5
& top1 & top5
& top1 & top5
& top1 & top5
& top1 & top5 \\

\midrule

\multicolumn{17}{l}{\textbf{Text-based Methods}} \\

BM25~\cite{robertson1994some}
& 60.23 & 82.55 & 59.14 & 82.14 & 65.79 & 86.87 & 54.05 & 77.84
& 59.83 & 84.94 & 7.25 & 20.73 & 52.95 & 72.90 & 11.84 & 39.35 \\

SBERT~\cite{reimers2019sentencebert}
& 22.77 & 41.83 & 21.82 & 41.12 & 25.74 & 48.54 & 27.43 & 51.33
& 27.99 & 52.25 & 4.02 & 17.29 & 17.72 & 36.67 & 13.06 & 41.24 \\

BGE-large~\cite{xiao2023cpack}
& 34.55 & 60.41 & 30.27 & 56.24 & 39.75 & 66.82 & 41.34 & 67.42
& 39.14 & 67.53 & 6.15 & 19.53 & 32.67 & 58.14 & 31.92 & 64.97 \\

BGE-M3~\cite{chen2024bgem3}
& 58.16 & 83.13 & 52.94 & 77.96 & 67.64 & 88.94 & 60.68 & 82.10
& 63.62 & 87.73 & 10.55 & 26.26 & \textbf{59.07} & \textbf{81.46} & 58.38 & 84.33 \\

NV-Embed-v2~\cite{lee2024nvembed}
& 42.92 & 72.71 & 40.84 & 66.32 & 52.23 & 80.30 & 49.41 & 76.13
& 47.12 & 78.74 & 5.47 & 21.73 & 41.86 & 68.30 & 42.17 & 72.70 \\

\textbf{BM25*}~\cite{robertson1994some}
& 61.07 & 84.98
& 60.30 & 84.96
& 66.87 & 89.36
& 58.43 & 84.14
& 62.18 & 87.13
& \textbf{25.92} & \textbf{47.20}
& 58.88 & 77.35
& 49.35 & 85.47 \\

\addlinespace[3pt]
\cdashline{1-17}

\multicolumn{17}{l}{\textbf{Multimodal Encoders}} \\

CLIP~\cite{radford2021clip}
& 11.14 & 29.32 & 12.39 & 31.77 & 19.53 & 45.69 & 19.52 & 44.44
& 16.22 & 42.71 & 4.64 & 18.91 & 10.46 & 27.36 & 14.28 & 44.86 \\

SigLip~\cite{zhai2023siglip}
& 13.08 & 32.36 & 17.52 & 40.69 & 25.69 & 51.69 & 24.85 & 53.15
& 22.70 & 50.85 & 5.53 & 19.56 & 13.98 & 33.56 & 15.62 & 46.20 \\

\addlinespace[3pt]
\cdashline{1-17}

\multicolumn{17}{l}{\textbf{Multimodal Large Language Models}} \\

VisRAG~\cite{yu2024visrag}
& 9.70 & 28.48 & 10.69 & 33.09 & 14.48 & 40.22 & 16.37 & 42.55
& 15.22 & 42.02 & 4.78 & 19.80 & 6.38 & 22.25 & 21.04 & 52.37 \\

VLM2Vec~\cite{jiang2024vlm2vec}
& 18.59 & 44.48 & 19.42 & 43.84 & 26.07 & 56.10 & 29.53 & 60.50
& 22.51 & 52.97 & 7.39 & 24.10 & 12.31 & 32.04 & 19.19 & 50.02 \\

GME~\cite{zhang2024gme}
& 60.57 & 88.08 & 52.96 & 79.08 & 65.97 & 89.61 & 66.78 & 89.55
& 57.92 & 85.16 & 15.33 & 35.72 & 45.09 & 72.60 & 61.11 & 89.37 \\

ColInternVL2~\cite{chen2024sdrag}
& 58.26 & 84.57 & 51.89 & 77.96 & 60.35 & 86.32 & 64.06 & 87.17
& 58.27 & 84.60 & 5.09 & 17.50 & 47.68 & 73.16 & 39.65 & 71.57 \\

ColPhi~\cite{chen2024sdrag}
& 65.42 & 89.00 & 56.06 & 81.43 & 65.02 & 88.96 & 67.83 & 89.65
& 62.15 & 88.17 & 8.46 & 25.95 & 48.83 & 74.82 & 25.28 & 56.49 \\

ColPali-v1.2~\cite{faysse2024colpali}
& 71.44 & 92.62 & 62.02 & 85.81 & 72.96 & 92.48 & 72.62 & 92.09
& 65.15 & 89.73 & 14.33 & 32.59 & 51.54 & 76.94 & 43.85 & 77.53 \\

ColQwen2-v0.1~\cite{faysse2024colpali}
& \textbf{75.04} & \textbf{94.34} & \textbf{65.18} & \textbf{88.24} & \textbf{78.63} & \textbf{95.77} & \textbf{77.81} & \textbf{93.69}
& \textbf{70.30} & \textbf{92.12} & 12.05 & 27.16 & 55.27 & 79.20 & \textbf{65.81} & 89.41 \\

\addlinespace[3pt]
\cdashline{1-17}

\multicolumn{17}{l}{\textbf{Finetuned on Multilingual Data}} \\

ColQwen2 (E)
& 67.25 & 90.60 & 57.10 & 82.29 & 71.99 & 93.18 & 72.01 & 91.39
& 60.26 & 86.57 & 10.15 & 26.70 & 44.50 & 71.12 & 61.54 & 87.70 \\

ColQwen2 (M)
& 69.77 & 92.48 & 59.77 & 85.39 & 72.71 & 92.97 & 72.70 & 92.16
& 64.37 & 89.10 & 11.67 & 28.07 & 50.44 & 76.83 & 63.91 & \textbf{89.98} \\

\addlinespace[4pt]

\midrule

\multicolumn{17}{l}{} \\

\addlinespace[2pt]

& \multicolumn{2}{c}{\textbf{Swedish}}
& \multicolumn{2}{c}{\textbf{Vietnamese}}
& \multicolumn{2}{c}{\textbf{Portuguese}}
& \multicolumn{2}{c}{\textbf{Finnish}}
& \multicolumn{2}{c}{\textbf{Czech}}
& \multicolumn{2}{c}{\textbf{Slovenian}}
& \multicolumn{2}{c}{\textbf{Danish}}
& \multicolumn{2}{c}{\textbf{Macro Average}} \\

\cmidrule(lr){2-3}\cmidrule(lr){4-5}\cmidrule(lr){6-7}\cmidrule(lr){8-9}
\cmidrule(lr){10-11}\cmidrule(lr){12-13}\cmidrule(lr){14-15}\cmidrule(lr){16-17}

Method
& top1 & top5
& top1 & top5
& top1 & top5
& top1 & top5
& top1 & top5
& top1 & top5
& top1 & top5
& top1 & top5 \\

\midrule

\multicolumn{17}{l}{\textbf{Text-based Methods}} \\

BM25~\cite{robertson1994some}
& 57.49 & 83.68 & 48.86 & 73.22 & 61.47 & 79.98 & 50.11 & 70.33
& 66.19 & 89.34 & 56.45 & 81.81 & 54.52 & 82.64 & 51.08 & 73.88 \\

SBERT~\cite{reimers2019sentencebert}
& 28.26 & 60.99 & 17.94 & 37.07 & 25.85 & 50.24 & 23.34 & 47.29
& 26.28 & 50.00 & 22.31 & 48.03 & 29.56 & 58.11 & 22.27 & 45.47 \\

BGE-large~\cite{xiao2023cpack}
& 42.18 & 74.69 & 23.94 & 48.97 & 38.53 & 66.08 & 31.58 & 57.97
& 33.97 & 61.94 & 35.30 & 63.89 & 35.58 & 71.02 & 33.12 & 60.37 \\

BGE-M3~\cite{chen2024bgem3}
& 65.25 & 89.33 & 44.93 & 68.82 & 60.07 & 82.16 & \textbf{56.90} & 77.19
& 65.87 & 90.22 & \textbf{65.05} & 88.53 & 64.42 & 88.38 & 56.90 & 79.77 \\

NV-Embed-v2~\cite{lee2024nvembed}
& 53.02 & 81.40 & 25.75 & 60.24 & 56.98 & 80.34 & 34.32 & 61.63
& 41.99 & 70.59 & 43.91 & 73.21 & 52.94 & 79.48 & 42.06 & 69.59 \\

\textbf{BM25*}~\cite{robertson1994some}
& 59.26 & 86.67
& \textbf{54.86} & \textbf{83.71}
& 62.68 & 85.56
& 56.22 & \textbf{80.32}
& \textbf{68.59} & \textbf{93.27}
& 61.20 & 85.93
& 53.37 & 85.65
& 57.28 & 82.78 \\

\addlinespace[3pt]
\cdashline{1-17}

\multicolumn{17}{l}{\textbf{Multimodal Encoders}} \\

CLIP~\cite{radford2021clip}
& 17.38 & 48.84 & 6.67 & 22.13 & 16.75 & 42.17 & 12.13 & 36.84
& 11.86 & 34.78 & 13.35 & 36.29 & 13.77 & 45.48 & 13.34 & 36.77 \\

SigLip~\cite{zhai2023siglip}
& 26.78 & 61.66 & 8.38 & 25.13 & 25.30 & 51.03 & 17.24 & 45.84
& 20.67 & 47.92 & 17.03 & 43.28 & 23.53 & 54.38 & 18.53 & 43.82 \\

\addlinespace[3pt]
\cdashline{1-17}

\multicolumn{17}{l}{\textbf{Multimodal Large Language Models}} \\

VisRAG~\cite{yu2024visrag}
& 14.93 & 49.30 & 5.53 & 18.56 & 12.68 & 38.29 & 9.76 & 34.78
& 9.46 & 34.13 & 8.69 & 34.50 & 13.77 & 46.34 & 11.57 & 35.78 \\

VLM2Vec~\cite{jiang2024vlm2vec}
& 25.98 & 62.17 & 8.22 & 25.39 & 23.73 & 53.46 & 16.17 & 44.47
& 21.07 & 50.56 & 15.59 & 44.09 & 25.39 & 58.68 & 19.41 & 46.86 \\

GME~\cite{zhang2024gme}
& 59.09 & 89.79 & 26.22 & 51.81 & 65.29 & 90.72 & 38.83 & 68.80
& 51.52 & 82.29 & 51.34 & 80.91 & 54.52 & 86.94 & 51.50 & 78.70 \\

ColInternVL2~\cite{chen2024sdrag}
& 61.16 & 90.51 & 25.75 & 54.19 & 62.32 & 86.95 & 46.22 & 72.85
& 55.29 & 85.90 & 54.75 & 83.96 & 60.11 & 90.10 & 50.06 & 76.49 \\

ColPhi~\cite{chen2024sdrag}
& 64.19 & 93.08 & 34.28 & 65.25 & 64.99 & 88.59 & 49.73 & 75.82
& 58.65 & 88.86 & 56.81 & 85.66 & 61.12 & 90.67 & 52.59  & 78.83 \\

ColPali-v1.2~\cite{faysse2024colpali}
& 65.37 & 92.11 & 35.32 & 66.60 & 76.03 & 92.96 & 42.11 & 73.07
& 62.34 & 91.27 & 55.82 & 86.29 & 62.41 & 90.10 & 56.89 & 82.15 \\

ColQwen2-v0.1~\cite{faysse2024colpali}
& \textbf{70.16} & \textbf{95.37} & 35.39 & 64.51 & \textbf{76.32} & \textbf{93.53} & 49.72 & 75.82
& 65.03 & 92.57 & 61.67 & 88.98 & \textbf{72.31} & \textbf{94.76} & \textbf{62.05} & \textbf{84.36} \\

\addlinespace[3pt]
\cdashline{1-17}

\multicolumn{17}{l}{\textbf{Finetuned on Multilingual Data}} \\

ColQwen2 (E)
& 62.30 & 91.37 & 24.50 & 53.29 & 49.44 & 80.19 & 40.88 & 68.43
& 52.87 & 84.12 & 49.44 & 80.19 & 63.92 & 92.22 & 52.54 & 78.62 \\

ColQwen2 (M)
& 66.12 & 94.05 & 27.10 & 55.05 & 71.86 & 92.09 & 40.88 & 70.04
& 61.32 & 89.63 & 58.43 & \textbf{89.35} & 64.52 & 93.22 & 57.04 & 82.03 \\

\bottomrule
\end{tabular}
}
\end{table*}

\subsection{BM25 Sensitivity to Representation Choices}
\label{app:bm25_sensitivity}

BM25 is highly sensitive to small representation changes: enabling or disabling a single step (e.g.,
segmentation for Japanese, morphology for Arabic) can change Top-5 accuracy by over 10 points even when OCR
is fixed (Table~\ref{tab:bm25_complete}). At the same time, the best configurations are typically simple
(one or two steps enabled), indicating that most gains come from lightweight, language-specific normalization
rather than complex pipelines (Table~\ref{tab:best_configs}).

\begin{table}[h]
\centering
\caption{Best configuration per language based on Top-5 accuracy. We report the optimal OCR model and linguistic processing features for each of the 15 languages tested.}
\label{tab:best_configs}
\small
\begin{tabular}{@{}lcccc@{}}
\toprule
\textbf{Language} & \textbf{Best OCR} & \textbf{Top-1 (\%)} & \textbf{Top-5 (\%)} & \textbf{Features} \\
\midrule
Spanish       & EasyOCR  & 61.07 & 84.98 & Stemming \\
Italian       & EasyOCR  & 60.30 & 84.96 & None \\
German        & Ministral 3B      & 66.87 & 89.36 & Stemming \\
French        & Ministral 3B      & 58.43 & 84.14 & None \\
Dutch         & EasyOCR  & 62.18 & 87.13 & None \\
Arabic        & Mistral OCR 3  & 25.92 & 47.20 & Morphology \\
Croatian      & Mistral OCR 3  & 58.88 & 77.35 & Lemmatization \\
Japanese      & Ministral 3B      & 49.35 & 85.47 & Adv. Tokenizers \\
Swedish       & Ministral 3B      & 59.26 & 86.67 & Stemming \\
Vietnamese    & Ministral 3B      & 54.86 & 83.71 & None \\
Portuguese    & Ministral 3B      & 62.68 & 85.56 & None \\
Finnish       & EasyOCR  & 56.22 & 80.32 & Stemming \\
Czech         & Ministral 3B      & 68.59 & 93.27 & Lemmatization \\
Slovenian     & EasyOCR  & 61.20 & 85.93 & Lemmatization \\
Danish        & EasyOCR  & 53.37 & 85.65 & Stemming \\
\midrule
\textbf{Average}  & ---      & 57.28 & 82.78 & --- \\
\bottomrule
\end{tabular}
\end{table}

\subsection{Multilingual Leaderboard Context}
\label{app:multilingual_leaderboard}

Table~\ref{tab:multilingual_full} provides the full multilingual leaderboard across text-based and multimodal
retrievers. The most important takeaway for interpretation is that the \emph{relative} standing of text-based
retrievers is strongly affected by representation choices (BM25 vs.\ BM25*), whereas multimodal methods are
invariant to OCR. This makes OCR/preprocessing a key confounder when multilingual comparisons mix modalities
(Table~\ref{tab:multilingual_full}). It is also worth noting that documents have different page-count distributions across languages, which should be considered when comparing results~\cite{chen2025visr}.

\begin{table*}[h]
\centering
\caption{Complete BM25 ablation study across all languages and OCR models on multilingual VisR-Bench. \checkmark=enabled, \texttimes=disabled. Top-1 / Top-5 accuracy (\%).}
\label{tab:bm25_complete}
\tiny
\setlength{\tabcolsep}{2.5pt}
\begin{tabular}{@{}llcccccccccccc@{}}
\toprule
& & \multicolumn{3}{c}{Adobe Text Extract} & \multicolumn{3}{c}{EasyOCR} & \multicolumn{3}{c}{Ministral 3B} & \multicolumn{3}{c}{Mistral OCR 3} \\
\cmidrule(lr){3-5} \cmidrule(lr){6-8} \cmidrule(lr){9-11} \cmidrule(lr){12-14}
Language & Config & MSLS & T-1 & T-5 & MSLS & T-1 & T-5 & MSLS & T-1 & T-5 & MSLS & T-1 & T-5 \\
\midrule
\multirow{2}{*}{Spanish} & 1 & \texttimes\checkmark\texttimes\texttimes & 57.65 & 82.00 & \texttimes\checkmark\texttimes\texttimes & 61.07 & 84.98 & \texttimes\checkmark\texttimes\texttimes & 56.98 & 82.92 & \texttimes\checkmark\texttimes\texttimes & 56.69 & 79.82 \\
                         & 2 & \texttimes\texttimes\texttimes\texttimes & 60.23 & 82.55 & \texttimes\texttimes\texttimes\texttimes & 58.49 & 81.70 & \texttimes\texttimes\texttimes\texttimes & 60.07 & 83.87 & \texttimes\texttimes\texttimes\texttimes & 59.03 & 80.18 \\
\midrule
\multirow{2}{*}{Italian} & 1 & \texttimes\checkmark\texttimes\texttimes & 56.57 & 80.70 & \texttimes\checkmark\texttimes\texttimes & 57.54 & 83.41 & \texttimes\checkmark\texttimes\texttimes & 55.56 & 81.33 & \texttimes\checkmark\texttimes\texttimes & 55.05 & 79.03 \\
                         & 2 & \texttimes\texttimes\texttimes\texttimes & 59.14 & 82.14 & \texttimes\texttimes\texttimes\texttimes & 60.30 & 84.96 & \texttimes\texttimes\texttimes\texttimes & 58.74 & 83.35 & \texttimes\texttimes\texttimes\texttimes & 57.42 & 80.09 \\
\midrule
\multirow{2}{*}{German} & 1 & \texttimes\checkmark\texttimes\texttimes & 66.09 & 87.82 & \texttimes\checkmark\texttimes\texttimes & 67.47 & 89.19 & \texttimes\checkmark\texttimes\texttimes & 66.87 & 89.36 & \texttimes\checkmark\texttimes\texttimes & 66.87 & 87.34 \\
                        & 2 & \texttimes\texttimes\texttimes\texttimes & 65.79 & 86.87 & \texttimes\texttimes\texttimes\texttimes & 62.90 & 85.82 & \texttimes\texttimes\texttimes\texttimes & 66.39 & 88.06 & \texttimes\texttimes\texttimes\texttimes & 66.29 & 85.89 \\
\midrule
\multirow{2}{*}{French} & 1 & \texttimes\checkmark\texttimes\texttimes & 52.72 & 78.02 & \texttimes\checkmark\texttimes\texttimes & 52.89 & 81.43 & \texttimes\checkmark\texttimes\texttimes & 56.02 & 83.68 & \texttimes\checkmark\texttimes\texttimes & 54.41 & 79.56 \\
                        & 2 & \texttimes\texttimes\texttimes\texttimes & 54.05 & 77.84 & \texttimes\texttimes\texttimes\texttimes & 53.87 & 80.64 & \texttimes\texttimes\texttimes\texttimes & 58.43 & 84.14 & \texttimes\texttimes\texttimes\texttimes & 56.89 & 79.76 \\

\midrule
\multirow{2}{*}{Dutch} & 1 & \texttimes\checkmark\texttimes\texttimes & 59.24 & 84.69 & \texttimes\checkmark\texttimes\texttimes & 62.30 & 86.73 & \texttimes\checkmark\texttimes\texttimes & 59.42 & 83.81 & \texttimes\checkmark\texttimes\texttimes & 58.77 & 82.65 \\
                       & 2 & \texttimes\texttimes\texttimes\texttimes & 59.83 & 84.94 & \texttimes\texttimes\texttimes\texttimes & 62.18 & 87.13 & \texttimes\texttimes\texttimes\texttimes & 59.96 & 84.10 & \texttimes\texttimes\texttimes\texttimes & 59.77 & 82.65 \\
\midrule
\multirow{2}{*}{Arabic} & 1 & \checkmark\texttimes\texttimes\texttimes & 7.84 & 22.59 & \checkmark\texttimes\texttimes\texttimes & 5.67 & 18.25 & \checkmark\texttimes\texttimes\texttimes & 19.11 & 40.84 & \checkmark\texttimes\texttimes\texttimes & 25.92 & 47.20 \\
                        & 2 & \texttimes\texttimes\texttimes\texttimes & 7.25 & 20.73 & \texttimes\texttimes\texttimes\texttimes & 4.78 & 16.98 & \texttimes\texttimes\texttimes\texttimes & 15.68 & 34.03 & \texttimes\texttimes\texttimes\texttimes & 19.08 & 37.88 \\
\midrule
\multirow{4}{*}{Croatian} & 1 & \texttimes\texttimes\checkmark\texttimes & 56.77 & 75.57 & \texttimes\texttimes\checkmark\texttimes & 54.47 & 76.49 & \texttimes\texttimes\checkmark\texttimes & 55.47 & 75.79 & \texttimes\texttimes\checkmark\texttimes & 58.88 & 77.35 \\
                          & 2 & \texttimes\checkmark\texttimes\texttimes & 52.95 & 72.90 & \texttimes\checkmark\texttimes\texttimes & 51.76 & 74.45 & \texttimes\checkmark\texttimes\texttimes & 51.98 & 73.34 & \texttimes\checkmark\texttimes\texttimes & 55.21 & 75.34 \\
                          & 3 & \texttimes\checkmark\checkmark\texttimes & 56.77 & 75.57 & \texttimes\checkmark\checkmark\texttimes & 54.47 & 76.49 & \texttimes\checkmark\checkmark\texttimes & 55.47 & 75.79 & \texttimes\checkmark\checkmark\texttimes & 58.88 & 77.35 \\
                          & 4 & \texttimes\texttimes\texttimes\texttimes & 52.95 & 72.90 & \texttimes\texttimes\texttimes\texttimes & 51.76 & 74.45 & \texttimes\texttimes\texttimes\texttimes & 51.98 & 73.34 & \texttimes\texttimes\texttimes\texttimes & 55.21 & 75.34 \\
\midrule
\multirow{2}{*}{Japanese} & 1 & \texttimes\texttimes\texttimes\checkmark & 48.72 & 80.55 & \texttimes\texttimes\texttimes\checkmark & 14.24 & 43.39 & \texttimes\texttimes\texttimes\checkmark & 49.35 & 85.47 & \texttimes\texttimes\texttimes\checkmark & 46.87 & 77.74 \\
                          & 2 & \texttimes\checkmark\texttimes\texttimes & 43.47 & 75.47 & \texttimes\checkmark\texttimes\texttimes & 10.50 & 37.76 & \texttimes\checkmark\texttimes\texttimes & 48.89 & 84.00 & \texttimes\checkmark\texttimes\texttimes & 44.94 & 76.14 \\
                          & 3 & \texttimes\texttimes\texttimes\texttimes & 11.84 & 39.35 & \texttimes\texttimes\texttimes\texttimes & 11.38 & 39.56 & \texttimes\texttimes\texttimes\texttimes & 11.93 & 39.82 & \texttimes\texttimes\texttimes\texttimes & 11.72 & 39.48 \\
\midrule
\multirow{2}{*}{Swedish} & 1 & \texttimes\checkmark\texttimes\texttimes & 58.75 & 85.36 & \texttimes\checkmark\texttimes\texttimes & 54.91 & 84.27 & \texttimes\checkmark\texttimes\texttimes & 59.26 & 86.67 & \texttimes\checkmark\texttimes\texttimes & 59.68 & 83.72 \\
                         & 2 & \texttimes\texttimes\texttimes\texttimes & 57.49 & 83.68 & \texttimes\texttimes\texttimes\texttimes & 54.32 & 83.30 & \texttimes\texttimes\texttimes\texttimes & 58.50 & 85.58 & \texttimes\texttimes\texttimes\texttimes & 58.75 & 82.45 \\
\midrule
\multirow{2}{*}{Vietnamese} & 1 & \texttimes\texttimes\texttimes\checkmark & 48.24 & 71.77 & \texttimes\texttimes\texttimes\checkmark & 18.05 & 40.80 & \texttimes\texttimes\texttimes\checkmark & 56.41 & 83.51 & \texttimes\texttimes\texttimes\checkmark & 57.60 & 81.33 \\
                            & 2 & \texttimes\texttimes\texttimes\texttimes & 48.86 & 73.22 & \texttimes\texttimes\texttimes\texttimes & 21.56 & 48.76 & \texttimes\texttimes\texttimes\texttimes & 54.86 & 83.71 & \texttimes\texttimes\texttimes\texttimes & 57.34 & 81.33 \\
\midrule
\multirow{2}{*}{Portuguese} & 1 & \texttimes\checkmark\texttimes\texttimes & 60.32 & 80.16 & \texttimes\checkmark\texttimes\texttimes & 54.79 & 81.01 & \texttimes\checkmark\texttimes\texttimes & 61.04 & 85.13 & \texttimes\checkmark\texttimes\texttimes & 61.77 & 83.25 \\
                            & 2 & \texttimes\texttimes\texttimes\texttimes & 61.47 & 79.98 & \texttimes\texttimes\texttimes\texttimes & 56.19 & 80.34 & \texttimes\texttimes\texttimes\texttimes & 62.68 & 85.56 & \texttimes\texttimes\texttimes\texttimes & 63.17 & 83.37 \\

\midrule
\multirow{4}{*}{Finnish} & 1 & \texttimes\texttimes\checkmark\texttimes & 49.12 & 68.57 & \texttimes\texttimes\checkmark\texttimes & 51.33 & 76.20 & \texttimes\texttimes\checkmark\texttimes & 51.03 & 75.13 & \texttimes\texttimes\checkmark\texttimes & 51.49 & 72.23 \\
                         & 2 & \texttimes\checkmark\texttimes\texttimes & 54.54 & 73.61 & \texttimes\checkmark\texttimes\texttimes & 56.22 & 80.32 & \texttimes\checkmark\texttimes\texttimes & 54.69 & 78.64 & \texttimes\checkmark\texttimes\texttimes & 56.83 & 75.36 \\
                         & 3 & \texttimes\checkmark\checkmark\texttimes & 49.12 & 68.57 & \texttimes\checkmark\checkmark\texttimes & 51.33 & 76.20 & \texttimes\checkmark\checkmark\texttimes & 51.03 & 75.13 & \texttimes\checkmark\checkmark\texttimes & 51.49 & 72.23 \\
                         & 4 & \texttimes\texttimes\texttimes\texttimes & 50.11 & 70.33 & \texttimes\texttimes\texttimes\texttimes & 51.18 & 75.13 & \texttimes\texttimes\texttimes\texttimes & 47.98 & 73.00 & \texttimes\texttimes\texttimes\texttimes & 51.18 & 72.01 \\
\midrule
\multirow{4}{*}{Czech} & 1 & \texttimes\texttimes\checkmark\texttimes & 68.91 & 92.15 & \texttimes\texttimes\checkmark\texttimes & 45.03 & 75.88 & \texttimes\texttimes\checkmark\texttimes & 68.59 & 93.27 & \texttimes\texttimes\checkmark\texttimes & 68.99 & 91.27 \\
                       & 2 & \texttimes\checkmark\texttimes\texttimes & 66.19 & 89.34 & \texttimes\checkmark\texttimes\texttimes & 44.63 & 73.32 & \texttimes\checkmark\texttimes\texttimes & 66.91 & 89.66 & \texttimes\checkmark\texttimes\texttimes & 67.31 & 88.62 \\
                       & 3 & \texttimes\checkmark\checkmark\texttimes & 68.91 & 92.15 & \texttimes\checkmark\checkmark\texttimes & 45.03 & 75.88 & \texttimes\checkmark\checkmark\texttimes & 68.59 & 93.27 & \texttimes\checkmark\checkmark\texttimes & 68.99 & 91.27 \\
                       & 4 & \texttimes\texttimes\texttimes\texttimes & 66.19 & 89.34 & \texttimes\texttimes\texttimes\texttimes & 44.63 & 73.32 & \texttimes\texttimes\texttimes\texttimes & 66.91 & 89.66 & \texttimes\texttimes\texttimes\texttimes & 67.31 & 88.62 \\
\midrule
\multirow{4}{*}{Slovenian} & 1 & \texttimes\texttimes\checkmark\texttimes & 60.57 & 84.14 & \texttimes\texttimes\checkmark\texttimes & 61.20 & 85.93 & \texttimes\texttimes\checkmark\texttimes & 62.99 & 84.23 & \texttimes\texttimes\checkmark\texttimes & 59.59 & 80.73 \\
                           & 2 & \texttimes\checkmark\texttimes\texttimes & 56.45 & 81.81 & \texttimes\checkmark\texttimes\texttimes & 53.32 & 82.89 & \texttimes\checkmark\texttimes\texttimes & 56.54 & 83.60 & \texttimes\checkmark\texttimes\texttimes & 54.84 & 79.03 \\
                           & 3 & \texttimes\checkmark\checkmark\texttimes & 60.57 & 84.14 & \texttimes\checkmark\checkmark\texttimes & 61.20 & 85.93 & \texttimes\checkmark\checkmark\texttimes & 62.99 & 84.23 & \texttimes\checkmark\checkmark\texttimes & 59.59 & 80.73 \\
                           & 4 & \texttimes\texttimes\texttimes\texttimes & 56.45 & 81.81 & \texttimes\texttimes\texttimes\texttimes & 53.32 & 82.89 & \texttimes\texttimes\texttimes\texttimes & 56.54 & 83.60 & \texttimes\texttimes\texttimes\texttimes & 54.84 & 79.03 \\
\midrule
\multirow{2}{*}{Danish} & 1 & \texttimes\checkmark\texttimes\texttimes & 55.81 & 84.07 & \texttimes\checkmark\texttimes\texttimes & 53.37 & 85.65 & \texttimes\checkmark\texttimes\texttimes & 57.39 & 85.08 & \texttimes\checkmark\texttimes\texttimes & 56.53 & 78.77 \\
                        & 2 & \texttimes\texttimes\texttimes\texttimes & 54.52 & 82.64 & \texttimes\texttimes\texttimes\texttimes & 52.80 & 84.51 & \texttimes\texttimes\texttimes\texttimes & 55.52 & 83.07 & \texttimes\texttimes\texttimes\texttimes & 55.81 & 77.33 \\

\bottomrule
\end{tabular}
\vspace{1mm}

\footnotesize
MSLS = Morphology/Stemming/Lemmatization/Segmentation. \checkmark  indicates enabled, \texttimes  indicates disabled. In the table we reuse the 'stemming' for Japanese to mark character level tokenization.
\end{table*}

\section{Retrieval Across Figure, Table, and Text Documents}
\label{app:modality_findings}

This section comments on patterns across Figure/Table/Text subsets and controlled OCR ablations across retrievers
(Table~\ref{tab:retrieval_combined_full}) that are not discussed in the main text.

\subsection{Dense Retrievers Also Benefit from Better Transcription}
\label{app:dense_ocr}

Although the main paper emphasizes BM25, the controlled OCR ablations show that dense retrievers also benefit
from improved transcription. In particular, changing only the transcription while keeping the retriever fixed
yields large gains for BGE-M3 and SBERT on figure-heavy pages (Table~\ref{tab:retrieval_combined_full}).
This indicates that the representation bottleneck is not specific to lexical matching: dense embedding models
are likewise constrained by missing or noisy text.

\subsection{Why Figure-Focused Gains Are Disproportionately Large}
\label{app:figure_gains}

The OCR ablations show substantially larger gains on figure-heavy pages than on text-only pages
(Table~\ref{tab:retrieval_combined_full}). A qualitative inspection suggests two main mechanisms:
(i) evidence is often contained in plot labels, legends, and axis annotations that are absent from default OCR, and
(ii) even coarse VLM-style transcriptions recover enough lexical anchors (numbers, variable names, captions) to improve
page discrimination (Table~\ref{tab:retrieval_combined_full}).

\begin{table*}[h]
\centering
\caption{\textbf{Retrieval accuracy (\%) on VisR-Bench across document types.}
\textbf{A:} Text-based retrieval with default OCR.
\textbf{B:} Controlled OCR ablation (retriever fixed; only transcription varies).
\textbf{C:} Multimodal retrievers operating directly on document images.
\textit{Macro Avg.} is the unweighted mean over Figure/Table/Text.}
\label{tab:retrieval_combined_full}
\resizebox{\textwidth}{!}{%
\begin{tabular}{llcccccccc}
\toprule
& & \multicolumn{2}{c}{Figure} & \multicolumn{2}{c}{Table} & \multicolumn{2}{c}{Text} & \multicolumn{2}{c}{Macro Avg.} \\
\cmidrule(lr){3-4}\cmidrule(lr){5-6}\cmidrule(lr){7-8}\cmidrule(lr){9-10}
Retriever & OCR / Transcription & top1 & top5 & top1 & top5 & top1 & top5 & top1 & top5 \\
\midrule

\multicolumn{10}{l}{\textbf{A. Text-Based Retrieval (Default OCR)}} \\

BM25~\cite{baeza1999modern}
& Adobe Text Extract
& 24.27 & 46.12 & 38.66 & 66.54 & 64.74 & 89.10 & 42.56 & 67.25 \\

SBERT~\cite{reimers2019sentencebert}
& Adobe Text Extract
& 25.24 & 49.27 & 26.31 & 52.68 & 49.96 & 76.97 & 33.84 & 59.64 \\

BGE-large~\cite{xiao2023cpack}
& Adobe Text Extract
& 31.55 & 56.07 & 40.36 & 70.14 & 57.00 & 82.68 & 42.97 & 69.63 \\

BGE-M3~\cite{chen2024bgem3}
& Adobe Text Extract
& 31.07 & 56.80 & 51.11 & 78.51 & 67.68 & 89.89 & 49.96 & 73.95 \\

NV-Embed-v2~\cite{lee2024nvembed}
& Adobe Text Extract
& 35.44 & 65.05 & 44.04 & 73.34 & 61.38 & 87.46 & 46.95 & 75.28 \\

\addlinespace[3pt]
\cdashline{1-10}

\multicolumn{10}{l}{\textbf{B. Controlled OCR Ablation (Retriever Fixed)}} \\

\multirow{4}{*}{BM25}
& EasyOCR$^{\ast}$
& 47.22 & 73.12 & 43.39 & 70.89 & 65.58 & 90.62 & 52.06 & 78.21 \\
& Mistral-3 OCR$^{\ast}$
& 45.52 & 71.91 & 41.61 & 68.12 & 66.34 & 91.38 & 51.16 & 77.14 \\
& Ministral~3B$^{\ast}$
& 49.88 & 77.24 & 41.51 & 69.34 & 65.03 & 90.10 & 52.14 & 78.89 \\

\addlinespace[4pt]

\multirow{4}{*}{SBERT}
& EasyOCR$^{\ast}$
& 39.95 & 69.73 & 29.58 & 58.46 & 49.95 & 77.86 & 39.83 & 68.68 \\
& Mistral-3 OCR$^{\ast}$
& 34.14 & 60.53 & 23.18 & 49.96 & 51.30 & 78.18 & 36.21 & 62.89 \\
& Ministral~3B$^{\ast}$
& 37.77 & 63.44 & 22.36 & 47.59 & 48.42 & 76.00 & 36.18 & 62.34 \\

\addlinespace[4pt]

\multirow{4}{*}{BGE-M3}
& EasyOCR$^{\ast}$
& 55.21 & 82.81 & 47.96 & 76.06 & 69.55 & 92.08 & 57.57 & 83.65 \\
& Mistral-3 OCR$^{\ast}$
& 51.57 & 79.42 & 45.42 & 74.92 & 69.10 & 91.66 & 55.36 & 82.00 \\
& Ministral~3B$^{\ast}$
& 59.32 & 87.41 & 48.06 & 77.51 & 67.88 & 91.83 & 58.42 & 85.58 \\

\addlinespace[3pt]
\cdashline{1-10}

\multicolumn{10}{l}{\textbf{C. Multimodal Retrieval (Reference)}} \\

CLIP~\cite{radford2021clip}
& --
& 33.90 & 61.74 & 24.68 & 47.59 & 39.47 & 70.21 & 32.68 & 59.85 \\

SigLip~\cite{zhai2023siglip}
& --
& 38.98 & 69.73 & 24.73 & 53.22 & 39.06 & 70.97 & 34.26 & 64.64 \\

VisRAG~\cite{yu2024visrag}
& --
& 31.96 & 66.83 & 19.82 & 48.53 & 31.00 & 61.49 & 27.59 & 58.95 \\

VLM2Vec~\cite{jiang2024vlm2vec}
& --
& 40.44 & 76.27 & 28.51 & 57.77 & 39.90 & 71.69 & 36.28 & 68.58 \\

GME~\cite{zhang2024gme}
& --
& 68.04 & 91.53 & 61.50 & 86.38 & 76.34 & 95.62 & 68.63 & 91.18 \\

Col-InternVL2~\cite{chen2024sdrag}
& --
& 68.28 & 90.31 & 63.85 & 86.36 & 79.19 & 96.45 & 70.44 & 91.04 \\

Col-Phi~\cite{chen2024sdrag}
& --
& 68.77 & 93.22 & 65.65 & 88.51 & 81.67 & 97.04 & 72.03 & 92.92 \\

ColPali-v1.2~\cite{faysse2024colpali}
& --
& 68.77 & 91.77 & 66.12 & 88.26 & 82.63 & 96.89 & 72.51 & 92.31 \\

ColQwen2-v0.1~\cite{faysse2024colpali}
& --
& \textbf{74.58} & \textbf{95.64} & \textbf{67.43} & \textbf{88.98} & \textbf{83.68} & \textbf{97.61} & \textbf{75.23} & \textbf{94.08} \\

\bottomrule
\end{tabular}
}
\vspace{3pt}
\caption*{\footnotesize
$^{\ast}$ OCR models are used strictly for transcription; retrieval configuration is unchanged.
Multimodal retrievers bypass OCR and are shown for contextual reference rather than direct comparison.}
\end{table*}

\subsection{Limits of Representation-Only Improvements}
\label{app:limits}

Even with the best OCR and normalization, text-based retrievers remain well below state-of-the-art multimodal
systems on visually grounded questions (Table~\ref{tab:retrieval_combined_full}). These cases correspond to
evidence that is not recoverable as text (e.g., spatial relations, graphical trends, non-textual encodings),
highlighting a clear boundary: representation improvements close much of the gap for text-bearing visual content,
but not for fundamentally visual reasoning (Table~\ref{tab:retrieval_combined_full}).

\subsection{Benchmarking Implication}
\label{app:benchmark_implication}

If changing only OCR yields double-digit gains for a fixed retriever (Table~\ref{tab:retrieval_combined_full}),
then benchmark gaps should not be interpreted as evidence of superior retrieval models. In such settings,
OCR and preprocessing should be treated as benchmark variables rather than hidden implementation details.

\newpage

\section{Example annotations}
In this section, we present example annotations produced by different text extraction methods, illustrating a progression in OCR quality: from limited extraction focused on figure captions, to broader text coverage, and finally to full image-level content descriptions. We opt for a simple prompt for the Ministral 3B - \emph{``Give me a markdown of what you see in the image. Reply only with the markdown content.''}. It would be an interesting area of exploration to optimize for lexical recall and query-agnostic retrieval effectiveness, e.g. prompting the model to provide concise, noun-heavy and non-redundant descriptions.

\example{Example 1}{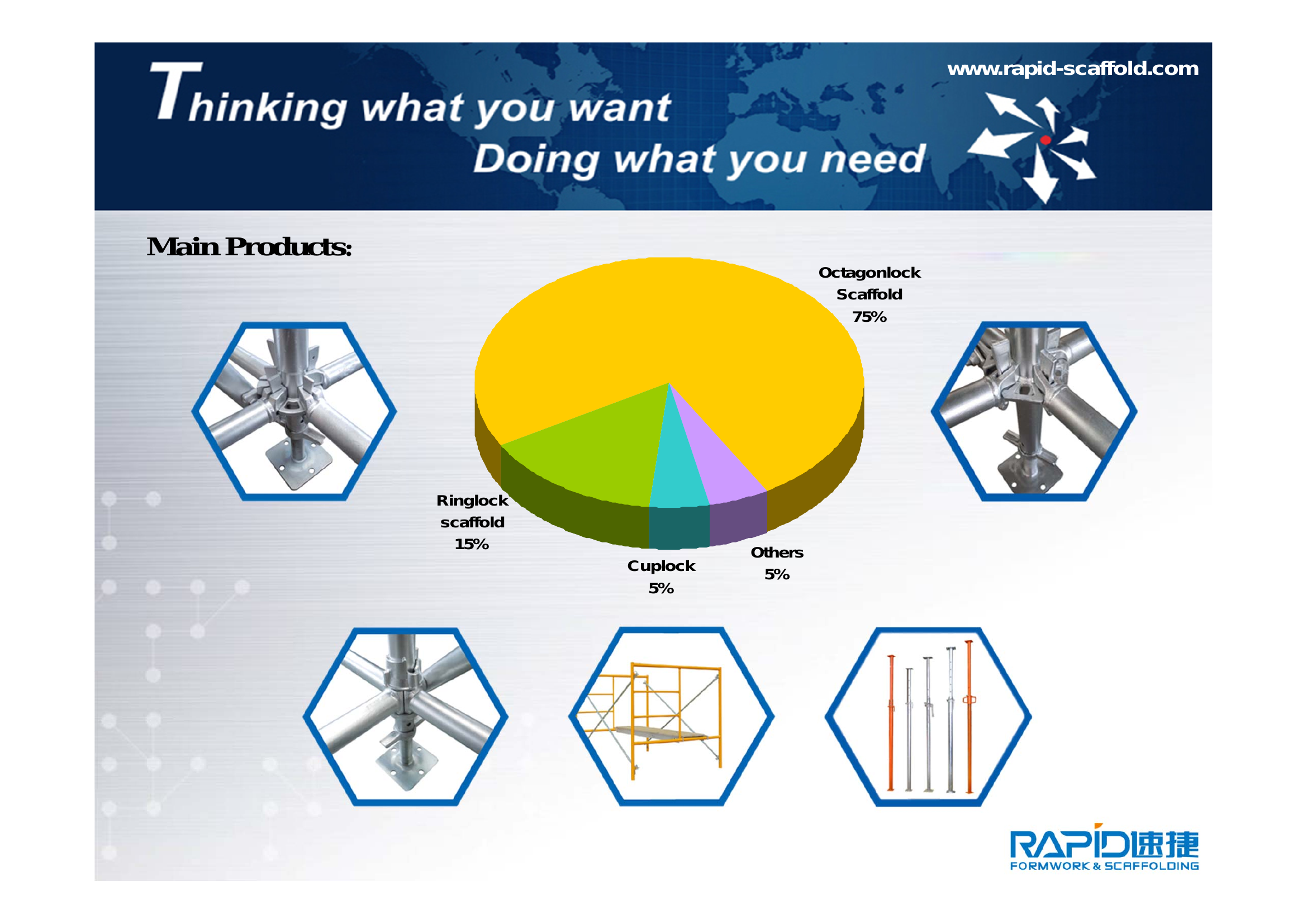}{doc3_page5}
\example{Example 2}{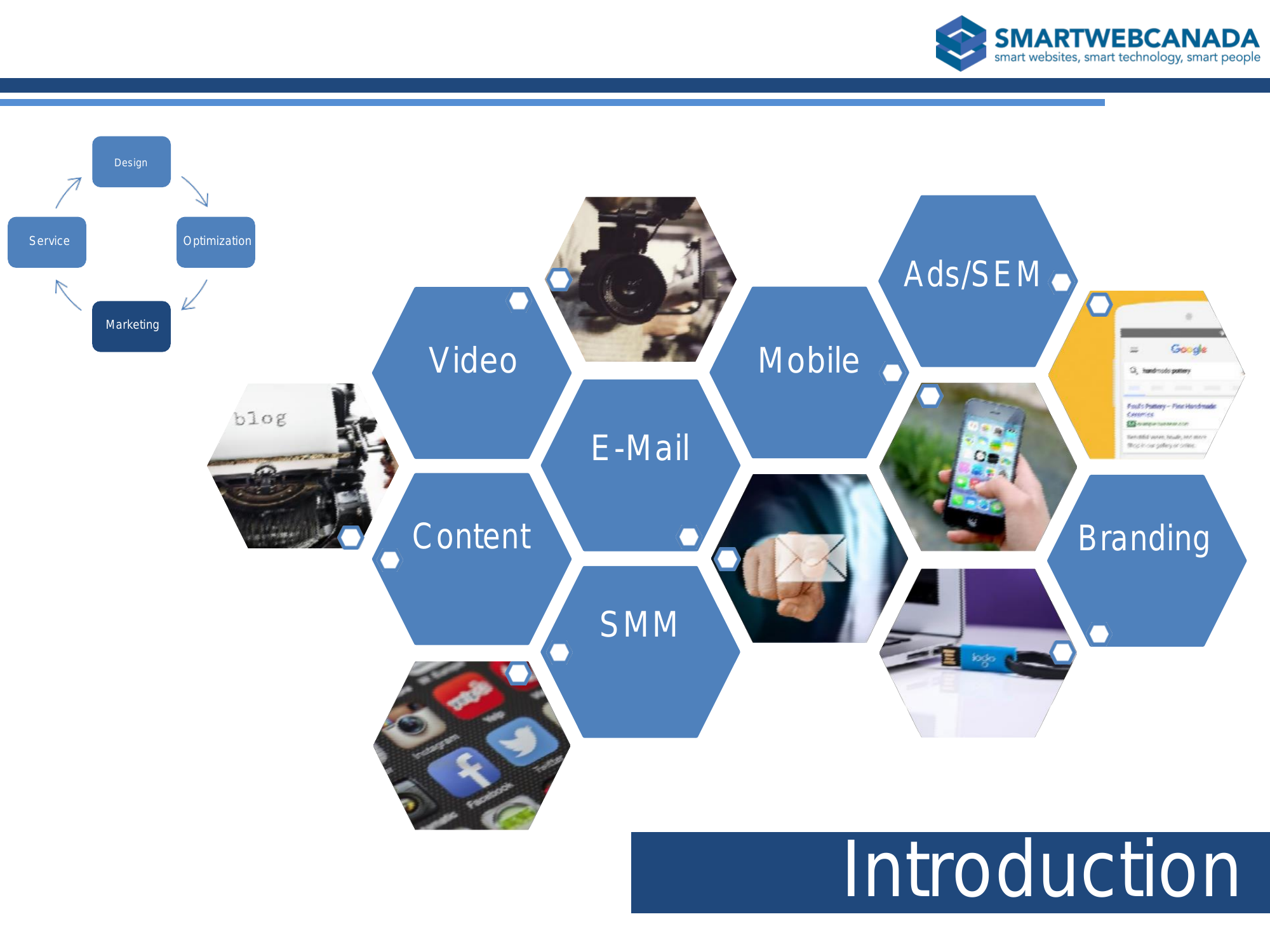}{doc5_page6}
\example{Example 3}{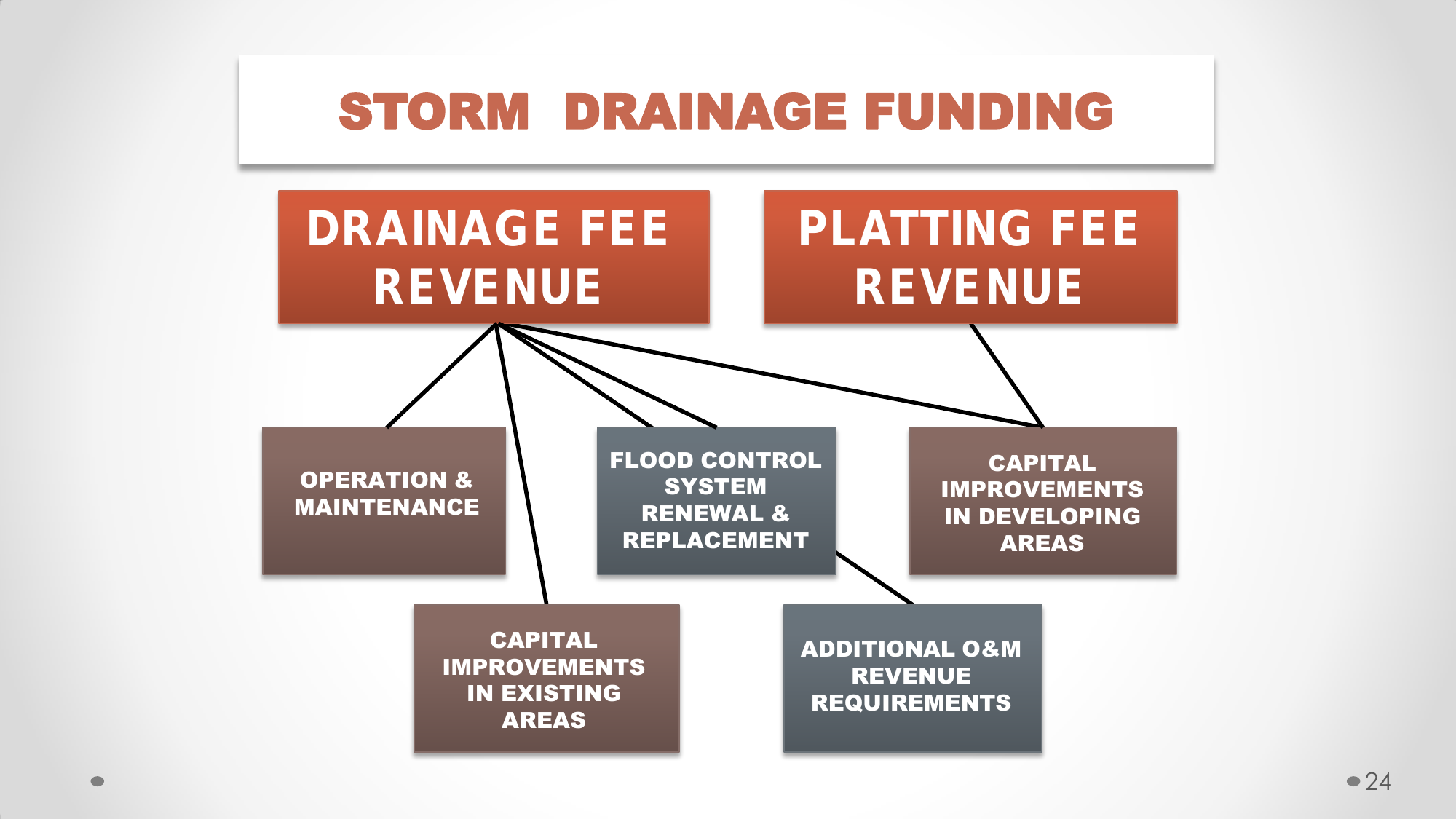}{doc0_page94}
\example{Example 4}{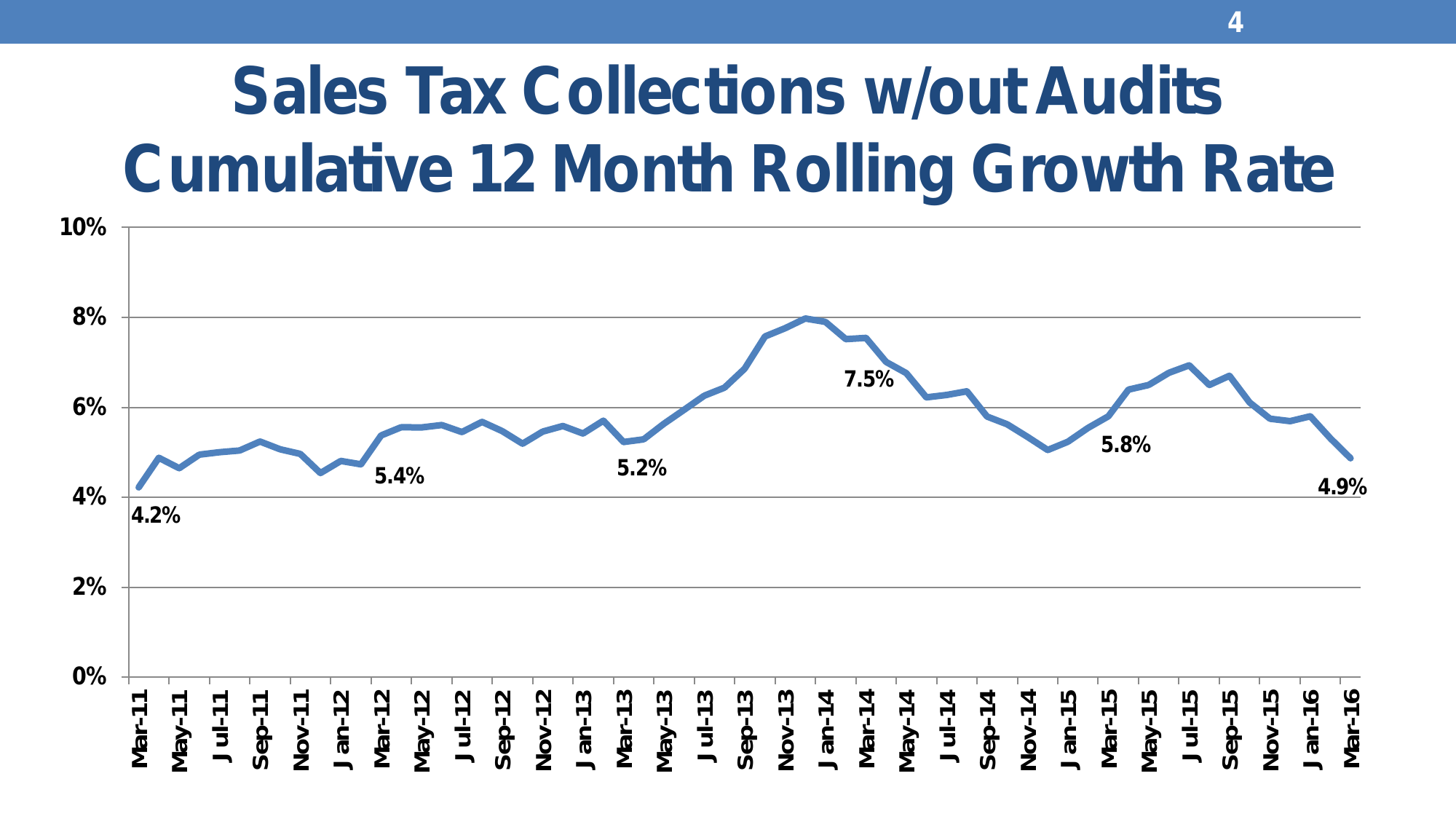}{doc0_page35}


\bibliographystyle{iclr2026_conference}
\end{document}